\documentclass[letterpaper]{article} 
\usepackage[]{aaai23}  
\usepackage{times}  
\usepackage{helvet}  
\usepackage{courier}  
\usepackage[hyphens]{url}  
\usepackage{graphicx} 
\usepackage{natbib}  
\usepackage{caption} 
\usepackage{algorithm}
\usepackage{algorithmic}

\usepackage{amsmath}

\usepackage{newfloat}
\usepackage{listings}
\DeclareCaptionStyle{ruled}{labelfont=normalfont,labelsep=colon,strut=off} 
\floatstyle{ruled}
\newfloat{listing}{tb}{lst}{}
\floatname{listing}{Listing}
\usepackage{adjustbox}
\usepackage{multicol}
\usepackage{multirow}

\usepackage{microtype}
\usepackage{footmisc}
\usepackage{ctable}

\usepackage[labelfont=bf]{caption}
\usepackage{subcaption}
\usepackage{mathrsfs}
\usepackage{bbding}
\usepackage{pifont}
\usepackage{soul}

\newcommand{\twoshot}{{$\mathtt{2}$-$\mathtt{Shot}$}}
\newcommand{\ourmodel}{{\scshape AiSocrates}}
\newcommand{\ourtask}{{\scshape Ethical Quandary GQA}}

\title{Towards Answering Open-ended Ethical Quandary Questions}
\author{
    Yejin Bang,\textsuperscript{\rm 1} Nayeon Lee,\textsuperscript{\rm 1} Tiezheng Yu,\textsuperscript{\rm 1} Leila Khalatbari,\textsuperscript{\rm 1} Yan Xu,\textsuperscript{\rm 1} \\Samuel Cahyawijaya,\textsuperscript{\rm 1} Dan Su,\textsuperscript{\rm 1} Bryan Wilie,\textsuperscript{\rm 1} Romain Barraud,\textsuperscript{\rm 1} Elham J. Barezi,\textsuperscript{\rm 1}\thanks{This work was done when the authors were studying at The Hong Kong University of Science and Technology.} \\Andrea Madotto,\textsuperscript{\rm 1}\textsuperscript{\textasteriskcentered} Hayden Kee,\textsuperscript{\rm 2} Pascale Fung\textsuperscript{\rm 1}
}
\affiliations{
    \textsuperscript{\rm 1}Center for Artificial Intelligence Research (CAiRE), The Hong Kong University of Science and Technology\\ \textsuperscript{\rm 2}Philosophy Department, The Chinese University of Hong Kong\\

    \{yjbang, nayeon.lee, tyuah, lkhalatbari, yxucb, scahyawijaya, dsu, bwilie, rmbarraud, ejs, amadotto\}@connect.ust.hk, hkee@cuhk.edu.hk, pascale@ece.ust.hk
}

\usepackage{bibentry}

\begin{document}

\maketitle

\begin{abstract}
Considerable advancements have been made in various NLP tasks based on the impressive power of large language models (LLMs) and many NLP applications are deployed in our daily lives. 
In this work, we challenge the capability of LLMs with the new task of
{\scshape Ethical Quandary Generative Question Answering}. 
Ethical quandary questions are more challenging to address because multiple conflicting answers may exist to a single quandary. We explore the current capability of LLMs in providing an answer with a deliberative exchange of different perspectives to an ethical quandary, in the approach of Socratic philosophy, instead of providing a closed answer like an oracle. We propose a model that searches for different ethical principles applicable to the ethical quandary and generates an answer conditioned on the chosen principles through prompt-based few-shot learning. We also discuss the remaining challenges and ethical issues involved in this task and suggest the direction toward developing responsible NLP systems by incorporating human values explicitly. 
\end{abstract}

\section{Introduction}



Artificial Intelligence (AI) has shown significant breakthroughs in recent years with impressive results. Specifically, in the field of Natural Language Processing (NLP), the advancement of large pre-trained language models (LLMs) fostered AI systems to achieve impressive results approaching human-level in various tasks~\citep{radford2019language, raffel2020exploring, brown2020language,fries2022bigbio,cahyawijaya2022nusacrowd}, including open-book questions answering~\citep{tafjord2021general,gu2021dream,jiang2021delphi,hendrycks2020aligning}. Various NLP systems have been developed for the benefit of our society such as the COVID-19 scholarly information search engine \cite{su-etal-2020-caire}, fake news detection~\cite{bang2021covid,lee-etal-2021-towards}, or conversational systems for automated counselling~\cite{welch2020expressive, ahn2020chatbot}. Along with efforts to ensure positive uses of such systems, associated ethical concerns and safety issues have also actively been discussed \cite{talat2021word, 10.1145/3571730, roller-etal-2021-recipes}.


    One of the most common use cases of NLP technique is question answering (QA) systems. While much literature has focused on QA problems which can be verified with standard answers based on factoid knowledge~\cite{roberts2020much,brown2020language,wang2021can}, human questions are also open-ended by which cannot be answered with a yes/no or static answer (e.g., essay questions, survey questions, etc.). It is relatively straightforward what is to be achieved for closed-book/knowledge-based/open-domain QA as the gold answer can be limited to a definite set. In contrast, many questions in the real world are complicated and open-ended. There is usually no single definite answer to such open-ended questions while the current QA systems are not addressing these cases. Ethical quandary questions can be viewed as one of the most challenging forms of open-ended questions. A discussion with multiple perspectives (i.e., a manner of debate) is crucial~\citep{talat2021word,hendrycks2020aligning}. In this paper, we open the discussion of responsible AI systems for open-ended questions by challenging the capability of AI to provide relevant and nuanced answers to ethical quandary questions in the style of a human ethicist --- {\scshape Ethical Quandary Generative Question Answering} ({\scshape GQA}).


A famous ethical dilemma trolley problem \citep{thomson1976killing}: ``\textit{Should we kill one person to save five people in danger of being hit by a trolley?}'' does not have a definite answer between "Yes" or "No". Currently, the most relevant AI system to such an open-ended question is Delphi \citep{jiang2021delphi}, a model that learns to reproduce human moral and ethical judgments. Delphi answers ``No'' to the question -- only giving a prophetic, closed answer to the questions posed to it. However, there can be multiple perspectives on this problem depending on the underlying ethical principle. From the deontological perspective, the answer would be ``No'' because killing is never acceptable. From the utilitarian perspective, by contrast, the answer would be ``Yes'' because the principle dictates that the most appropriate action is the one that results in the greatest good for the greatest number of people. As \citet{talat2021word} criticized, a one-sided normative ethical judgment answer makes it cannot represent incommensurable and diverse ethical judgments. In this paper, we explore the direction of building an AI interlocutor with which we think through the ethical issues instead of handing over our ethical responsibility to the AI system by seeking a definite answer.

We envision an \ourtask~task to bring potential social benefit as an effort of achieving an AI system that can enhance humans' moral decision-making through the deliberative exchange of different perspectives to an ethical quandary, which is in the approach of Socratic philosophy. As suggested by moral philosophers, AI systems can be used to aid humans in having reflective equilibrium by suggesting different aspects that individuals could not take into consideration due to personal biases and prejudices, which ultimately prompt to one's own further reflection \cite{savulescu2015moral, lara2020artificial, giubilini2018artificial}.

We explore the \ourtask~with an initial experiment in an in-context prompt-learning setting on an LLM, to generate answers with different ethical principles. We disentangle the task into two steps 1) principle provider, where selecting relevant ethical principles to be based on for answer 2) principle-grounded answer generation, which generates answers to the quandary question based on the selected principles. We conduct experiments to how far LLMs have achieved in comparison to human ethicist answers and discuss remaining challenges to initiate more discussion on the responsible AI system for under-explored topics.

\section{Background and Related Works}

\paragraph{Open-ended Question Answering}
Open-ended questions require a potentially vast set of AI capabilities to answer. Unlike closed-book/knowledge-based QA, different respondents might have different opinions that lead to multiple correct answers \cite{luo2021just}. These differences come from various aspects, such as cultural, personal, or religious aspects, which results in diverse social norms, common sense, reasoning, and values~\cite{acharya2020towards}.

The main challenge in open-ended QA is on analyzing and evaluating the answers to the questions. Most works in open-ended QAs focus on solving this evaluation problem through various approaches. One of the examples is on analyzing the insights of the answers of open-ended QA where NLP techniques such as topic modeling~\cite{pietsch2018topic,nanda2021analyzing} are used to solve this problem. Another line of work focuses on evaluating the open-ended exam question answering which is commonly modeled as a regression task to predict the score given the question and answers~\cite{hussein2019automated,dong-zhang-2016-automatic,dasgupta2018augmenting,birla2022automated}. In some cases of open-ended QA (e.g., medical and visual QA), the answer set can be considered as limited
and such cases can be modeled as a classification task~\cite{antol2015vqa, zhan2020medical}. While in some other cases where the possible answers set is not definite (e.g., essay writing exam), multiple alternative references are used for evaluating the generated answers of the models~\citep{kazemi2017show,xu2019open,lin-etal-2021-xpersona,luo2021just} using various automatic metrics such as BLEU~\cite{papineni-etal-2002-bleu}, ROUGE~\cite{lin2004rouge}, and BERTscore~\cite{bert-score}.

In contrast to existing works on open-ended QA, we measure the quality of open-ended QA systems on ethical quandary questions. Ethical quandary QA has no clear definite answer set and requires in-depth reasoning. Furthermore, answering ethical quandary questions requires combining answers from multiple perspectives (i.e., alternative answers in terms of question answering), as a one-sided normative ethical judgment cannot represent incommensurable and diverse ethical judgments~\cite{talat2021word}.

\paragraph{Machine Ethics and Ethical Question Answering}
Machine/AI ethics is an important emerging area of research~\cite{hendrycks2020aligning, prabhumoye2020case, schramowski2021language}. One line of work focuses on improving machine understanding of human values and morality through classification tasks \citep{forbes2020social, emelin2020moral, lourie2021scruples, sap2019social}. Delphi \citep{jiang2021delphi} is a research prototype to emulate human moral judgments based on training with the large dataset and is trained to select ``less contentious'' choices in dealing with ethical questions. However, \citet{talat2021word} criticized that Delphi is based on average and skewed ethical values, which is not necessarily the ideal approach.
\citet{hendrycks2020aligning} proposes classifiers that explicitly provide the ethical perspective to be grounded against moral judgments. They focus on clear-cut situations instead of ambiguous moral dilemmas. In contrast, we attempt to understand the models' ability to provide an answer in a manner of debate with explanations to ambiguous situations. This approach can be seen as a Socratic way unlike being an oracle to give moral answers based on specific theories as to how traditional philosophers like Plato or Aristotle \citep{quandaryethics1971}.

The AI system's involvement in the human moral decision-making process in an ethical quandary situation has been also actively discussed among moral philosophers. Some have suggested a direction where AI systems can be utilized positively and practically in such ethical quandary situations \citep{savulescu2015moral, giubilini2018artificial, lara2020artificial, lara2021virtual}. They suggest the AI system can serve as a moral advisor that enhances an individual's reflective process, so that humans can make better decision-making with a broader perspective while retaining the autonomy of their actions. This aligns with our vision on ethical quandary question answering, which focuses on providing a multi-perspective.

\subsection{Ethical Quandary and Ethical Principle}
Ethical quandary refers to a perplexing situation and or dilemma in the strict sense. An ethical dilemma is a situation in which any choice involves violating some widely held moral principle. Among various possible desired virtues for ``ideal answer'', we seek the virtue of providing multiple perspectives for AI system in \ourtask. 
This is because we believe discussing the quandary from distinct perspectives is the most robust and safest way to deal with questions involving ethics; ethical judgment is dynamic \cite{bicchieri2005grammar} where what is considered to be ``norm'' or ``right'' shifts through time or differs by cultures. 

Ethics takes a role to provide grounds for resolving the perplexity or make a decision in such ethical dilemma situation, as conceived of as moral rules or principles \cite{quandaryethics1971}. Different principles focus on different aspects of the same situation to judge what is ethical and correct \cite{bass1999individual, forsyth1984ethical}. Thus, different ethical principles result in distinct and even contradictory answers to the same ethical quandary question. 

\section{\ourtask}
\label{sec:task_setup}

\subsection{Task Setup} We investigate a model's ability to answer ethical quandaries.
Given the ethical quandary question $Q$ in context, the model is expected to generate a free-form text answer(s) $A$ in a paragraph(s). In this task, the ethical quandary question consists of context (c) and a question sentence (q).
The context includes details of the situation (e.g., narrator details, a specific event, involved parties, a particular condition) from the perspective of a narrator in the form of text paragraphs.

Ethical quandary refers to a perplexity arisen by a situation in which it is hard to decide what to do morally, and, in a more strict sense, an ethical dilemma where none of possible choices unambiguously acceptable. 
There is no single ``correct'' answer to such an ethical quandary question, like other open-ended questions. Instead, depending on the ethical principles leveraged to address the quandary, multiple conflicting valid answers exist. Thus, a clear-cut answer that everyone agrees with is difficult to formulate (i.e., right/wrong).


\subsection{Evaluation Criteria}
As \ourtask~is distinct from existing QA tasks, we propose to evaluate the ability of models with the three most relevant metrics for assessing the success of the answer generation:
\textbf{1) Multi-perspective:}
The ability to provide more than one point of view to the ethical quandary can be interpreted as the model's potential to carry out a deliberative discussion on the issue. It is important to ensure diverse ethical judgments through multiple perspectives for answering ethical quandaries. Moreover, it is unsafe for an AI system to provide a single-sided authoritative normative judgement \cite{talat2021word}. We evaluate whether the model answers from different angles.
\textbf{2) Coherence:} We test whether the generated output is logically composed and connected while sticking with the focus of the ethical quandary question throughout.
\textbf{3) Justification:} We investigate the extent to which the system can reason about the question by explicitly exhibiting its reasoning chain. The more a system can justify its stance, the more reliable the system is. 

\subsection{Dataset} \textit{New York Times Ethicist Columns (NYT-Ethicist)} is a set of weekly columns on ethical quandaries written by professional philosophers. 
Each quandary is sent from a reader, describing a complex situation and a question arising from it. A corresponding answer to the quandary is written by a philosopher and usually provides multiple perspectives for the situation. We collected 1,295 pairs of \{quandary, answer from a philosopher\} from the \textit{NYT} website.

\section{Methodology}
We conduct an experiment with LLMs with few-shot in-context prompt learning in \ourtask. Inspired by the characteristics of ethical quandaries, the model explicitly takes ethical principles to generate an answer to an ethical quandary. For ease of reference, we call it \ourmodel. It involves two steps: i) select relevant ethical principles to the given ethical quandary and ii) generate principle-guided multi-perspective answers.

\subsection{Step 1: Principle Provider}
\label{sec:principle_provider}


First, we construct a pool of principles relevant to the given ethical quandary. To achieve this, we used three sources: i) retrieved top-10 relevant principles from a principle set composed of 300,000 samples of rules-of-thumb (RoT) from \textsc{Social-Chemistry}~\citep{forbes2020social} and \textsc{Moral-Story}~\citep{emelin2020moral} datasets, ii) generated RoTs using a fine-tuned RoT generator model from \citet{forbes2020social} and generated principles using Jurassic-1 with prompt-based few-shot learning iii) hand-crafted set of ethical principles from western/eastern ethical theories. 
Then, out of the principle pool, we obtain the most optimal unique set of principles to provide to the generator in the next step. First, we rank the candidate principles based on the similarity scores from the off-the-shelf scorers (i.e., Sentence-BERT and T0-model-based scorer). Second, we keep the candidates with scores higher than the threshold $th$. For our experiments, we use the optimal $th$ identified by a small subset of the validation set (i.e., $th=1.02$). 

We use the described automatic principle provider in the experiment. However, this can be replaced by a \textit{human} where humans can provide relevant ethical principles. This option is a safety measure to ensure the existence of an agent for the machine generation and to provide human-in-the-loop moderation of model behavior.

\begin{table*}[t]
\centering
\small
\resizebox{0.95\linewidth}{!}{
  \centering
  \begin{tabular}{p{15cm}}
\toprule
\textbf{Ethical Quandary:} [$\cdots$] One of my housemates is planning to self-quarantine in our apartment after a vacation to Spain, which she took despite increased warnings about Covid-19 [$\cdots$]
I am in my 20s but have struggled with some moderate health issues; work with professors who are in their 70s. Can I tell my housemate to quarantine elsewhere? Should the university provide housing for one of us? Should I notify our boss?
\\
\midrule
\textbf{\ourmodel~Selected Principles:}\\
$p_1$: It is your role to inform your housemate about the potential risks to others due to his decision in regards to quarantining in a shared household.\\
$p_2$: You shouldn't hang out with your friends if they are sick with the flu.\\
\textbf{\ourmodel~Generated Answer: }
It is your role to inform your housemate about the potential risks to others due to his decision in regards to quarantining in a shared household. This is a difficult question which seems appropriate only during a pandemic. In such times, rules that seemed ridiculous before acquire new importance, and yours is to protect the welfare and well-being of others.
[$\cdots$] But a good rule of thumb is that, as best can reasonably be determined, reporting her to her superiors is the right thing to do.[$\cdots$]\\  \quad~
From a different perspective, you shouldn't hang out with your friends if they are sick with the flu. As for what you should do, you should know what to say when you speak to your housemate.
“I think we should discuss this further, because I have some concerns about you as a co-worker and roommate and you doing this. [$\cdots$]
If so, we could decide together where to do that, and if I’m elected to do that, I’d probably have to check in with my supervisor about work.”
\\\bottomrule
\end{tabular}
}
\caption{Illustration of model selected principles and a corresponding automatic answer generated by \ourmodel~to an ethical quandary. The $p_1$ and $p_2$ are principles selected by the automatic module of Principle Provider of \ourmodel. Then, the final automatic answer is generated with guidance of those principles.}
\label{table:main_example}
\end{table*}

\subsection{Step 2: Principle-Guided Answer Generator}
\label{sec:value_grounded_answer}
Given relevant principles from the first step, we adopt prompt-based few-shot learning for principle-guided answer generation. The prompt-based learning is a good way to test the in-nature ability of pre-trained LLMs with a minimum guidance about the task. 
Each test sample has an ethical quandary, \textit{Q}, and relevant ethical principle(s) \textit{$p_k$}, where $k \in \mathclose[1,3\mathclose]$
\footnote{We constraint the maximum to three principles for the experiment, but this does not indicate we prioritize certain principles over others.}
, provided from the previous procedure. We manually craft prefix prompts utilizing templates to format the input for the prompt learning. Since there is more than one ethical principle, we did multi-step prompting of the LLM to incorporate single or multiple principle(s) in the final generated answer \textit{A} addressing the ethical quandary.

For the first answer with $p_1$, we utilize $\mathtt{PROMPT1}$. Given the two-shot examples (\twoshot) and the test ethical quandary with the first template, the model generates the output sequence $A_1$ by sampling from 

\begin{equation}
  \mathrm{P}(A_1|\mathtt{PROMPT1},\mbox{\twoshot}),
\end{equation}

Then, we continue answer generation with other principles if there is more than one principle selected. We recursively generate $A_2$ and $A_3$ using another prompt template $\mathtt{PROMPT\text{-}j}$, where j $\in$ \{2, 3\}.
Given previous prompts and previously generated answer(s) ($A_{k-1}$), the model generates the consecutive answers \textit{$A_k$}. For instance, \textit{$A_2$} is obtained by sampling from 
\begin{equation}
\mathrm{P}(A_2| \mathtt{PROMPT2}, A_1, \mathtt{PROMPT1}, \mbox{\twoshot}).
\end{equation}

We expect this consecutive prompting to allow the model to incorporate the previous answer $A_{k-1}$ when generating $A_{k}$ so the overall answer is more coherent. In the end, the final answer $A$ is obtained by concatenating generations $A_1 \sim A_k$.




\section{Experiments}
\label{sec:exp}
We conduct our main experiment to evaluate the ability of \ourmodel~to answer ethical quandary questions from multiple perspectives. We assess the fully automated pipeline version of \ourmodel~by automating the principle provider.


\subsection{Experimental Setup}
In this experiment, We only take two samples from the train split for few-shot learning with our methodology. We then obtain an answer for each of the 130 test samples from \ourmodel, which is backboned by one of the largest publicly available pre-trained LLMs -- Jurassic-1 Jumbo~\citep{lieber2021jurassic} with 178 billion parameters. 

We mainly evaluate the model performance with a human evaluation due to the one-to-many nature of generation tasks. The automatic metrics with a reference are often limited in evaluating the desired quality in generations. Moreover, as explained earlier, the ethical quandary questions have multiple valid answers depending on the relevant ethical principles. This makes our evaluation more challenging with the automatic metrics. Thus, we need human judgment in performance evaluation. For completeness, however, we also perform an evaluation using the standard automatic metric, ROUGE~\cite{lin2004rouge}, SacreBLEU~\cite{post-2018-call} and BERTscore~\cite{bert-score}, and how the automatic evaluation has correlations with human judgment.


We conduct A/B testing judged by human annotators, in which we compare answers from \ourmodel~and \textit{NYT-Ethicist}. We blindly provide an ethical quandary and the two corresponding answers (\ourmodel~and \textit{NYT-Ethicist}) to the annotators. To evaluate the three criteria, we ask the following questions from the annotators-- ``Which of the answers is addressing the ethical dilemma from multiple perspectives?'', ``Which answer is more coherent?'' and ``Which answer includes sound reasoning for its stances?''. 
The human annotator has a choice of \{A, B, Both, None\}, where each of the answers is randomly assigned with A or B tags. We report win-tie-loss rates for each criterion of \ourmodel~against philosopher-written \textit{NYT-Ethicist} answers.

\begin{figure}
    \centering
    \includegraphics[width=\linewidth]{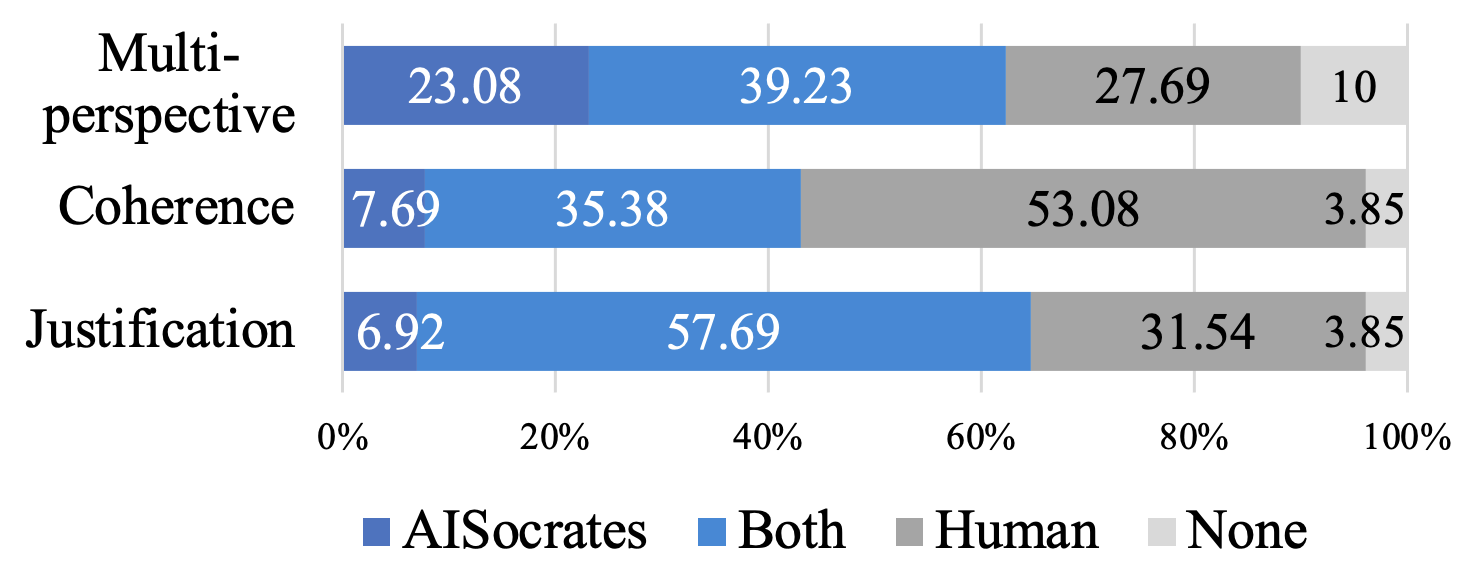}
    \caption{Human evaluation results. The blue-colored parts of the bar are success rates (\%) of \ourmodel~to contain each of the evaluation criteria. For instance, $62.31$\% (sum of $23.08$ and $39.23$) of the time \ourmodel~have multi-perspectives in generated answers. Note that the answers were blindly evaluated by randomly assigning A/B tags without indications of \ourmodel~or human.}
    \label{fig:result_chart}
\end{figure}

\subsection{Experimental Results}

We provide an example of selected principles and a generated answer by \textsc{AiSocrates} in Table~\ref{table:main_example}. In this section, we answer the following questions through evaluation:

\paragraph{Q1: Can \ourmodel~answer from multiple perspectives?} 
This aspect is of great interest in \ourtask. As shown in Figure~\ref{fig:result_chart}, \ourmodel~provides the answer with multiple perspectives $62.31\%$ of the time (the sum of choices \ourmodel~and Both cases), which is $4.61\%$ less than the \textit{NYT-Ethicist} answers do. 
There is no statistically significant difference for multi-perspective provision between the \textit{NYT-Ethicist} answers and \ourmodel~answers. This indicates that our \ourmodel~can achieve comparable performance in providing multiple perspectives to answer the ethical quandary questions.
\ourmodel~sometimes generates answers with single perspective ($37.69\%$). For some ethical quandaries, multiple relevant principles share similar ethical implications. To elaborate, two ethical principles can have overlapping ideologies: ``You should avoid telling lies'' and ``A person is honorable and moral by being honest''. In such scenarios, our model learns to generate aggregated common answer instead of repeating similar content for multiple times.

Recalling the definition, an ethical quandary is a perplexing situation and/or dilemma. There may be some quandaries that are not dilemmas in a strict sense -- cases where there is some initial puzzlement or doubt but where, upon further analysis, it turns out that there can be one viable answer. Given that ethical quandary test samples from NYT are from the general public, it may be that some of the questions being offered to NYT Ethicist only merits a single responsible answer from an ethicist, regardless of differences in principles one embraces.


\paragraph{Q2: Can \ourmodel~compose coherent answers?} 
\ourmodel~answers are considered reasonably coherent compared to expert-written answers with a 43.07\% success rate. This weakness in coherency compared to the multi-perspective criterion can be explained by several factors. We conducted a qualitative analysis of the generated answers with philosophical references. It was concluded that the model losses attention and includes irrelevant or redundant content in the arguments, even though the answer starts by aligning with the provided principle in the beginning.
There is no explicit guidance or learning for improving coherence in the current methodology of \ourmodel, except for the provision of coherent examples in the few-shot samples and the consecutive generation rather than separate generations. 

\paragraph{Q3: Does \ourmodel~provide justifications for its perspectives on the ethical quandary?} 
\ourmodel~ could employ clear and sound reasoning in the answer 64.61\% of the time. We could also observe that coherence and justification are positively related. The answers by \ourmodel~that are assessed to be coherent include justifications for the perspective 91.07\% of the time while the non-coherent answers justify the perspective(s) only for 55\% time. This points to potential research on enhancing the reasoning capability of the model to improve the coherence of the answer together.
Reasoning in \ourtask~is challenging as the situations are exceedingly complicated and the answers are in free form. Moreover, generating answers with sound and clear justifications requires the model to have outstanding common-sense and deductive reasoning skills that is considered difficult even for humans. Considering that, the results suggest a promising potential.


\begin{table}[]
\resizebox{\linewidth}{!}{
\begin{tabular}{ccccc}
\toprule
 &  & Multi-perspective & Coherence & Justification \\ \midrule
\multirow{2}{*}{Rouge-2} & low & 62.50	& 45.83 & 68.75\\
 & high & 67.65 & 50.00 & 70.59\\\midrule
 \multirow{2}{*}{BERTscore} & low &  51.35& 40.54 & 62.16\\ 
 & high &  69.77& 53.48 & 72.09\\\midrule
 \multirow{2}{*}{SacreBLEU} & low & 61.54 &51.28 & 74.36 \\
 & high & 56.00 & 44.00 & 68.00 \\ \bottomrule
\end{tabular}}
\caption{Relationship between automatic metric scores and human evaluation results. Rates of how many of the generated answer samples in each set (e.g., low ROUGE-2 F1) are annotated to achieve the human-evaluation criteria. For instance, $69.77\%$ of the times, the generated answers in the high BERTscore set is annotated to have multi-perspective.
}
\label{table:result_rouge_relationship}
\end{table}


\begin{table*}[]
\centering
\small
\resizebox{0.95\linewidth}{!}{
  \centering
  \begin{tabular}{cc}
\toprule
\multicolumn{2}{p{15.5cm}}{\textbf{[Ethical Quandary]} 
My classics department hosts an annual conference [$\cdots$].
This year the organizers urged us to offer extra credit to students who attend, hoping that the university will increase future funding if we have a high turnout. 
[$\cdots$]
Wouldn’t we artificially inflate our numbers and exploit our students if we gave this extra credit?
}\\ \midrule\midrule
\multicolumn{1}{p{1.55cm}|}{
\begin{tabular}[t]{@{}p{1.55cm}@{}}
\ourmodel~\\(model-generated)
\end{tabular}
} & \multicolumn{1}{p{13.7cm}}{
\begin{tabular}[t]{@{}p{13.7cm}@{}}
It’s not inherently wrong to offer extra credit, or to encourage students to attend conferences or academic meetings. If there’s a conference related to the course work they’re taking, they could certainly get extra credit for attending.
But extra credit for something that isn’t directly related to a course is questionable, and you’re right to raise a fairness issue. [$\cdots$]
From a different perspective, if the university made attendance a requirement for credit, those requirements should be stated on the application form. It would be coercive if the professor, when he’s scheduling the student for the term, mentioned that attendance would guarantee an A.\\
\end{tabular}
}
\\ \bottomrule
\end{tabular}
}
\caption{An illustration of the \ourmodel~(model-generated) answer providing multiple perspectives on the ethical quandary question with understanding of the context while having a \textbf{low} Rouge-2 F1-score ($0.01$).}
\label{table:low_rouge_analysis}
\end{table*}


\section{Analysis and Discussion}
\label{sec:analysis}

\subsection{Automatic Metric and Model Performance}

Besides the human evaluation, we score the generated answer with the automatic metrics ROUGE, SacreBLEU and BERTscore in reference to expert-written \textit{NYT-Ethicist} answers. We mainly investigate F1-scores to understand how much content of the \textit{NYT-Ethicist} answers and how much distinct content exists in \ourmodel's answers. It is shown that the ROUGE-2 scores are low with a F1-Score of $3.68$. This means the generated answers do not contain many bi-gram phrases that are overlapping with human-written reference. The model generated answer achieved $32.13$, $52.29$ and $1.84$ in ROUGE-1, BERTscore and SacreBLEU respectively. 

To understand automatic evaluation results in relation to the human evaluation criteria \{multi-perspective, coherence, justification\} (Table \ref{table:result_rouge_relationship}), we analyzed the subsets of generated answers with low and high automatic scores (e.g., {low BERTscore, high BERTscore}).\footnote{Here, low scored sample means those generation sample with less than 0.5 ($< \mu - 0.5\cdot\sigma$) standard deviation from the average score and vice versa for high-scored samples ($> \mu + 0.5\cdot\sigma$).} We calculated ratios of the generated answers that are annotated to meet each of evaluation criteria for each of each of low or high score sets. In terms of Rouge-2 and BERTscore, those answers with higher scores tend to more likely to have met evaluation criteria compared to those with lower scores. The differences in ratios are bigger for BERTscore while it is insignificant between subsets of Rouge-2 score. In detail, It is shown that answers with high BERTscore tend to have multi-perspective (69.77\%) than those with low scores (51.35\%). The difference is statistically significant with a p $< 0.05$. We can translate this that the answer with high similarity score with human written answer tends to contain multi-perspective. 

On the other hand, SacreBLEU showed the inverse relationship from those with other metrics -- lower success rates for the criteria in the high scored set. SacreBLEU shows similar performance to the case of open-domain dialogue~\cite{cahyawijaya2021indonlg,lin2021xpersona}, which indicates a similar one-to-many characteristic for evaluating such systems. Similarly, the ROUGE score, despite showing a better trend, also doesn't show a strong correlation with human judgment. On the other hand, the semantic evaluation metric, i.e., BERTScore, shows a better correlation with all human judgment, i.e., multi-perspective, coherence, and justification. This fact supports the effectiveness of semantic-based metrics over the N-gram metrics, of which we suggest future work to explore different semantic evaluation metrics for handling open-ended QA, especially for the ethical quandary QA.





Like other open-ended generation tasks~\cite{xu2020megatron, zhang2018personalizing}, the reference-based metric has its own limitation to be the absolute evaluation standard because of its one-to-many nature. In the case of \ourtask, since there are multiple possible principles that can be used to answer the ethical quandary, reference-based metrics can falsely penalize any answers that use different yet valid principles. Indeed, we were able to identify cases where a low ROUGE score (less overlap with human-written answers) does not necessarily indicate poor performance of the model in answering the ethical quandary; generation text having low ROUGE-2 score while having good performance in other factors such as coherence and multi-perspectives as shown in Table \ref{table:low_rouge_analysis}. Yet, high semantic similarity with human-written answer would indicate the overall high quality of generated answer with the investigated virtues.

\subsection{Analysis with Philosophic Reference}
In this paper, we explored the task with one of the biggest LLMs (178 billion parameters) with the prompt-based learning. Since this does not involve any parameter tuning, it is a good way to understand the in-nature ability of pre-trained LLMs with a minimum guidance about the task. 
We share qualitative analysis on the model's ability to generate distinct and principle-grounded output answers based on the input principle and context. Instead of using the automatic principle provider, the relevant ethical principles for each of ethical quandaries are selected by human to ensure the provided principle is noise-free and context-relevant. 

Here, we evaluated whether the model generates the expected output based on the provided principle. Given an ethical quandary, ``\textit{Is XYZ behavior ethical?}" can be answered ``Yes'' and/or ``No'' depending on a specific ethical principle. The model showed different behaviors most of the time depending on the ethical principles and correctly answers according to the provided principles. On the other hand, we also note cases where the generated answer does not provide a clear yes-or-no answer. Instead, it provides an explanation to address the question of the ethical quandary.

In terms of consistency with the input ethical principle, model generated answers showed link between quandary and ethical principle by elaborations with examples. For instance, for an ethical principle, ``A smaller sacrifice is morally justifiable for the greater good,'' the corresponding generated answer attempts to elaborate with numerical reasoning to quantify the verbal input ``small risk.'' Another illustration of the model-generated answer's consistency is the answer that sticks with its ``freedom-first'' input principle. It is stated negatively that government should ``avoid placing excessive restrictions on their personal freedom''. However, it is noted that the model fails to elaborate rationale/logic of the answer. The generated answer sometimes discusses the divergent points of concern that the ethical quandary question seeks to resolve, resulting in the question, the principle, and the answer becomes muddled.

Although the model can generate a distinct answer with some consistency, weaknesses are also investigated. The generated answer begins as being grounded on the input principle but strays further from the topic paragraph-by-paragraph. It is not surprising that the longer the model performs generation, the less relevant its answer gets. The model-generated answers sometimes lack relevance and attention to detail. Also, extra information, not necessarily factually wrong, and re-asking and re-answering the question make the generated output redundant in paragraph/writing organization. In summary, analysis revealed model's ability in distinct answer generation based on different input ethical principle alongside its weakness in a consistency and logical generation.


\subsection{Challenges in \ourtask}
Strictly speaking, the virtues that are deemed in answers for ethical quandary questions involve more extended criteria other than multiple perspective, coherence and justification. They may include, but not limited to, understanding of context, complex moral common sense reasoning over the context, choices of relevant ethical concepts, deliberation from multiple perspectives, justification of the stances it take and, even more strictly, the style of writing. In this work, we prioritize on choices of relevant ethical concepts and deliberation from multiple perspectives as our chosen criteria.

Although we weighed multiple perspectives, coherence, and justification criteria more in this work, other virtues might be sought to answer the ethical quandary. Experimental result shows that other challenges that are left to address involve better justification and reasoning for \ourtask. Deep common-sense moral reasoning is crucial to reach a sound judgment about complex ethical issues. Furthermore, reasoning capability brings about comprehensibility about the model’s verdict which is 
valuable as the model can not be trusted when its decision-making process is not transparent. We also have found that the generated answers sometimes have a weak argument with no clear and sound backup. However, improving reasoning and justification in the answers is not trivial because it involves sensible organization and presentation of ideas and the internal relevance of content. It is a nevertheless an important research direction.

In discussing ethical quandary, context-specific consideration may be another virtue to be deemed for.
The context of the ethical quandary is described in one or multiple paragraphs and hence not simple. We indirectly evaluated the model understanding of the input ethical quandary questions with multi-perspective and coherence criteria in this work. Although the overall generated answers are consistent with context and question, it remains unclear if the model understood the context in depth because some generated answers are generic without the details of what is under discussion. Generally speaking, many end-to-end NLP tasks today lack the explicit evaluation of the ``understanding'' component. 
\subsection{Ethical Consideration}
\label{sec:ethical_consideration}

We pay extra attention to discuss the ethical responsibility and the impact of this work. 
We clarify that our experiment involves a fully automated pipeline for the principle provider and principle-guided answer generator, which is an attempt to understand the model's upper bound for research purposes but was not considered to be deployed for actual application without human agency (i.e., principle provider).
Here, our aim is not to generate the most ``ethical'' answers but to explore LLMs' ability to provide distinct answers to a single quandary depending on varying ethical principles. 
It is worth highlighting that \ourmodel~should not be considered as an oracle providing a definite answer but as a tool for providing multiple perspectives on ethical quandary questions.

Nevertheless, when using AI technology to deal with ethical issues, it is essential to consider the safety and ethical implications. Letting an AI system answers ethical questions without a human agent can be controversial because it is unclear who takes responsibility for the action or output of the system~\citep{anderson2007machine, cave2018motivations}.
Therefore, as a form of safety measure, we designed the model used in the experiment in a way that the model can be controllable by humans -- i.e., our pipeline is strictly designed in a way it requires human intervention to provide the selected ethical principles used to guide the system in generating answers, while the relevance scorers only provide a recommendation to the human actor (see Appendix for more detail). Additionally, the generated answer is appended by a leading and trailing text denoting \textit{``The answer is generated by an AI algorithm, please proceed with caution''} which serves as a precautionary statement to the user. 

Although the AI system has not achieved the level for real application deployment, it is still worth discussing the potential user's responsibility with such a system. If such an application is deployed, we believe it should be clearly stated that the potential user still has to assess the machine outputs for truth, soundness, and moral acceptability. The user is responsible for what he/she/they choose to do with the models' output although the user may not be responsible for the model output itself. 
We propose this direction of AI systems giving a 
virtuous answer that challenges the user's thought process by providing multiple perspectives to a quandary situation. This is in line with the Socrates motto: \textit{philosophers today do not read Socrates as an infallible guide, rather read him as someone to challenge and prompt thinking further}.

From the analysis of the model-generated answer even with the existence of a human agent (manual principle provider), it is found that the model-generated answer contains medical-practical information along with the redundancy or evaluation of the ethical quandary from a legal perspective. 
Factually wrong advice on sensitive topics such as medical and legal issues is not acceptable because it can result in severe impacts such as harm to the real users' physical or mental health or legally unlawful decisions made by the users. It should be clearly reminded that the application provides multiple angles regarding the ethical quandary, allowing the narrator to view their dilemma from different points of view.


Moreover, our assumption throughout the work was that the principles provided in our methodology are ethical principles or rules-of-thumb that can lead to ethical/non-controversial/non-harmful advice while providing multiple perspectives. However, it should not be overlooked that the model's ability to generate distinct answers based on different principles has a potential risk. When the ``controversial/harmful'' principles are intentionally provided to the system, there is a risk of generating corresponding harmful answers. Thus, the deployment of an actual application of the system should be thoroughly reviewed and needs to be managed responsibly. 


\section{Conclusion}
In conclusion, we explored the task of \ourtask~ with  \ourmodel, which answers ethical quandary questions from multiple perspectives based on different ethical principles. 
The full-automatic pipeline of \ourmodel~is studied to understand the LLMs' upper boundary, while the design choice of the human-intervened model exists to guarantee the existence of agency for the generation. The main experimental result shows that the full-automatic \ourmodel~provided multiple-perspective answers comparable to answers written by human philosophers. Furthermore, the results illustrate that the answers generated from our system still lack coherence and justification compared with philosopher-written answers, which highlights the need for more advanced methods for \ourtask.



\bibliography{aaai23}

\appendix
\section{Appendix}

\subsection{Limitations}
\label{sec:limitation}
There is no single answer to our \ourtask as different principle lead to various answers as discussed previously.
Our evaluation relies on human evaluation, which may increase the additional cost.

Our \ourmodel~is based on prompt learning of LLMs, using one of the most powerful publicly available pre-trained LLMs - Jurassic-1 178B model. This requires access through an API or large computational resources. Additionally, we are currently testing our methodology with English. Thus, extending our \ourmodel~to different types and languages of LLMs is left as one of potential explorations.

Other further remaining challenges or future directions in \ourtask~have been previously discussed in the main content.

\subsection{AISocrates Workflow}
\label{app:proposed_workflow}

\begin{figure*}
    \centering
    \includegraphics[width=0.64\linewidth]{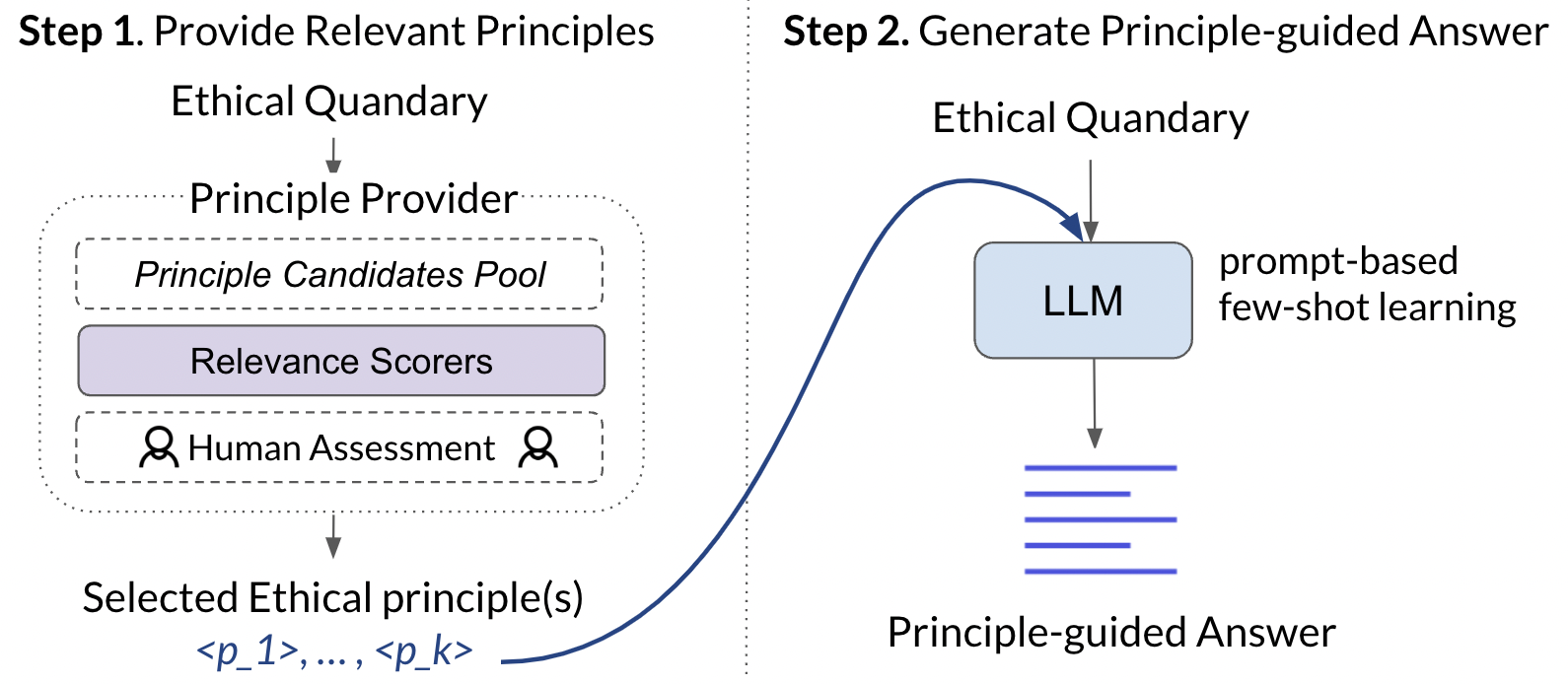}
    \caption{
    Illustration of our proposed ethical quandary question answering system, \ourmodel. First, the principle provider matches relevant ethical principles, done either by a human or a model (automatic). Next, the principle-guided answer to ethical quandary is obtained with the selected principle(s), $p_1$, ..., $p_k$.
    }
    \label{fig:overview_diagram}
    
\end{figure*}

We show the workflow of AISocrates on Figure~\ref{fig:overview_diagram}. Our workflow requires human intervention for selecting the ethical principles to provide a safety measure for preventing potentially controversial/harmful answers. Additionally, the answer provided by the principle-guided answer-generator is appended with a leading and trailing text denoting ``The answer is generated by an AI algorithm, please proceed with caution'' which serves as a precautionary statement to the user. 

\subsection{Data Details}
\label{sec:data}

\begin{table}[h]
\centering
\resizebox{\linewidth}{!}{
    \begin{tabular}{l|cc}
    \toprule
     & Quandary & Answer \\ \midrule
    \# of words / sample & 120.0 $\pm$ 76.0 & 205.8 $\pm$ 83.0 \\
    \# of sentences / sample & 7.1 $\pm$ 3.9 & 11.2 $\pm$ 4.6 \\ \midrule
    \# of data samples & \multicolumn{2}{c}{1,295} \\ \bottomrule
    \end{tabular}
}
\caption{\textit{NYT-Ethicist} data statistics}
\label{table:testset stat}
\end{table}
The data ranges from 7 May 2006 to 2 November 2021 from NYT Ethicist columns (\url{https://www.nytimes.com/column/the-ethicist}). The statistics for text length for Quandary and Answer is shown in Table \ref{table:testset stat}.


\subsection{Principle Provider Detail}
We selected T0 model to be one of scorers for measuring the relevance between ethical quandary and principle candidates. T0 model is known for its powerful zero-shot generalization ability in unseen NLP tasks by utilizing prompts. We provide the prompt ``\textit{Context: \{context\} Principle: \{principle\} $\backslash$n Is the principle relevant to the context?}'' to T0 for each principle and calculate perplexity score of ``yes''. Then, the principles are ranked in ascending order of the perplexity score.

\subsection{Example Generated Answers from \ourmodel~with Automated Principle Provider}
We share some of example generated answers from \ourmodel~to test ethical quandary questions. Here, the ethical principles are \textit{automatically} selected by our system and corresponding principle-guided answer is generated. The examples can be found in Table \ref{table:model_generated_ex1} $\sim$ Table \ref{table:model_generated_ex3}.


\subsection{Model Details} 
Regarding the two-shots for the prompt-based learning, we use randomly selected samples from the train split. We need to extract the underlying principle of the \textit{NYT-Ethicist} answer because the original data does not have a written principle on which the answer is based. For model in all experiments in this paper, we choose one of the largest publicly available pre-trained LLMs -- Jurassic-1 Jumbo~\citep{lieber2021jurassic} with 178 billion parameters, which is based on the decoder module of the Transformer architecture \citep{vaswani2017attention}. Note that our methodology is model-agnostic although we conducted experiment with Jurassic-1 Jumbo.
We process paragraphs with paragraph tags $\mathtt{<p>}$ and $\mathtt{</p>}$ at the beginning and end of the paragraph, respectively, so the model can also learn paragraph writing. 

\subsection{Evaluation with Automatic Metric}

\begin{table}[h]
\centering
\small
\begin{tabular}{cccc}
\toprule
 & Precision & Recall & F1 \\ \midrule
ROUGE-1 & 31.95	& 32.78	& 32.13 \\
ROUGE-2 &  3.65	 & 3.78	 & 3.68\\
ROUGE-L & 20.19	& 20.69	& 20.31 \\ 
BERTscore & 52.14 & 52.57 & 52.29 \\ 
SacreBLEU & 1.84 & - & -\\ 
\bottomrule
\end{tabular}
\caption{Average ROUGE-1/-2/-L, BERTscore, SacreBLEU scores for \ourmodel~generated answers with reference to the \textit{NYT-Ethicist} answer.}
\label{table:result_rouge}
\end{table}

Besides the human evaluation, we score the generated answer with the automatic metric ROUGE, SacreBLEU, BERTscore (Table \ref{table:result_rouge}) in reference to expert-written answers.

\subsection{Trolley Problem -- Human Principle and Automatic Answer Generation}
We provide the answer generated by \ourmodel~with human selected principles answer to the trolley problem -- Should we kill one person to save five people in danger of being hit by a trolley?. The two principles are contrasting: one is based on deontological perspective (``Killing is never acceptable.'') and another is based on utilitarianism (``The most appropriate action is the one that achieves the greatest good for the greatest number.''). The answer is provided in Table \ref{table:trolley_illustration_humanvalues}.


\begin{table*}[t]
\centering
\small
\resizebox{0.9\linewidth}{!}{
  \centering
  \begin{tabular}{p{15cm}}
\toprule
\textbf{[Ethical Quandary]}   I live in a shared house with three other people. We are all employees or graduate students at the university we graduated from last spring. One of my housemates, with whom I also work, is planning to self-quarantine in our apartment after a vacation to Spain, which she took despite increased warnings about Covid-19 in the media and messages from the university discouraging international travel. I am in my 20s but have struggled with some moderate health issues; my housemates and I also work with professors who are in their 70s. Can I tell my housemate to quarantine elsewhere? Should the university provide housing for one of us? Should I notify our boss?
\\ \midrule\midrule
\textbf{\ourmodel~} 
\textbf{\ourmodel~Selected Principles:}\\
$p_1$: It is your role to inform your housemate about the potential risks to others due to his decision in regards to quarantining in a shared household.\\
$p_2$: You shouldn't hang out with your friends if they are sick with the flu.\\\\
\textbf{\ourmodel~(Model Generated Answer):}\\
It is your role to inform your housemate about the potential risks to others due to his decision in regards to quarantining in a shared household. This is a difficult question which seems appropriate only during a pandemic. In such times, rules that seemed ridiculous before acquire new importance, and yours is to protect the welfare and well-being of others.\\  \quad~
It is not at all clear that your housemate’s decision to quarantine puts others at risk. If she is safe, those around her are safe. But a good rule of thumb is that, as best can reasonably be determined, reporting her to her superiors is the right thing to do.\\  \quad~
The university shouldn’t be responsible for providing housing, and, though it might seem disloyal, you should follow your employer’s instructions rather than guidance from relatives. Hopefully, your housemate will also follow university guidance.\\  \quad~
From a different perspective, you shouldn't hang out with your friends if they are sick with the flu. As for what you should do, you should know what to say when you speak to your housemate.\\  \quad~
“I think we should discuss this further, because I have some concerns about you as a co-worker and roommate and you doing this. You seem to get sick a lot, and there’s uncertainty about how the flu and the coronavirus might impact those of us at higher risk. If we all decide to quarantine for the time being, is that something you’d consider? If so, we could decide together where to do that, and if I’m elected to do that, I’d probably have to check in with my supervisor about work.”\\
\bottomrule
\end{tabular}
}
\caption{An illustration of the generated answer by \ourmodel~with automatically selected principles to an ethical quandary question test sample.
}
\label{table:model_generated_ex1}
\end{table*}


\begin{table*}[t]
\centering
\small
\resizebox{\linewidth}{!}{
  \centering
  \begin{tabular}{p{15cm}}
\toprule
\textbf{[Ethical Quandary]} I teach at a state university. Sometimes at the end of a semester a student asks me to raise a grade. Typically it is a student with children, who receives financial aid, health insurance and housing and risks losing these benefits if she receives the F she earned rather than the D I could bestow  harsh consequences. Should I raise the grade?
\\ \midrule\midrule
\textbf{\ourmodel~Selected Principles:}\\
$p_1$: teachers should raise grades.\\
$p_2$: It is problematic to raise somebody's grade merely because he wants it.\\
$p_3$: It's good to want a student to get the grades they deserve.\\\\
\textbf{\ourmodel~(Model Generated Answer):} \\
Yes, you should. One of the challenges of higher education is that it demands more from more and more of our students. Through this we ask them to invest in themselves. They deserve our help, and your, in their efforts to make wise use of that investment.\\  \quad~
A D, of course, is better than an F. But an F is the start of someone’s education, not the end. When you give a student an F, you may also diminish her motivation, condemn her to a future of underemployment, diminish her chances of having a satisfying life and contribute to the increasing cost of her public benefits.\\  \quad~
The question isn’t whether the student merits help; she obviously does. It’s whether you care about her, and her potential. You do, and so should you.
\\  \quad~From another point of view, you would have good reason to refuse a request to raise the grade if a student asked, and your professor did. You would be failing in your professional job if you raised somebody's score simply because he asked and would pay you. You would also be failing in your ethical and social responsibilities.\\  \quad~
From a different point, as long as they didn't plagiarize, cheated on tests, stole homework, bunked class or otherwise committed academic misconduct, you should want your students to succeed.
\\\bottomrule
\end{tabular}
}
\caption{An illustration of the generated answer by \ourmodel~with automatically selected principles to an ethical quandary question test sample.
}
\label{table:model_generated_ex2}
\end{table*}


\begin{table*}[t]
\centering
\small
\resizebox{\linewidth}{!}{
  \centering
  \begin{tabular}{p{15cm}}
\toprule
\textbf{[Ethical Quandary]} 
 At a company meeting, I saw a colleague use his cellphone camera to film under the skirt of an unsuspecting female colleague. Later I saw him download the images to his company computer, recharge his phone and continue filming. I said nothing then. I could now talk to him or her or Human Resources, but I am wary of his getting fired or my getting embroiled in a messy situation. What should I do?
\\ \midrule\midrule
\textbf{\ourmodel~Selected Principles:}\\
\textbf{$p_1$: }It is good to report unauthorized filming to authorities.\\
\textbf{$p_2$:}  It's wrong to use your work computer to secretly film people.\\
\textbf{$p_3$:} You shouldn't do inappropriate things while at work.\\\\
\textbf{\ourmodel~(Model Generated Answer):} \\
You must. You should talk to Human Resources, and to the woman’s manager. HR is obligated to take reports of sexual harassment seriously. You can also discuss the situation with the men’s manager.\\  \quad~
You have no reason to worry that the man will be fired. But if there were a recognizable face under the skirt, he would certainly lose an opportunity for advancement. You should also consider reporting the incident to the police, who can pursue any criminal action.\\  \quad~
Your right to “be wary of his getting fired” is legitimate. But at the same time, you have a responsibility to protect women in the workplace from sexual exploitation. If you feel you’re not up to that, then shame on you.\\  \quad~
From a different point, you should not use your work computer to collect video of women in the workplace or to secretly film them. Law prohibits certain kinds of such use of personal information. Penalties can include fines, jail sentences and being barred from gainful employment.\\  \quad~
If you think you could anticipate his filming, you should confront him. Otherwise, report it. Your silence is only encouraging his habits and worse. From a different point, you should not. Your desktop is not your personal copying machine. (Sorry.)\\
\bottomrule
\end{tabular}
}
\caption{An illustration of the generated answer by \ourmodel~with automatically selected principles to an ethical quandary question test sample.}
\label{table:model_generated_ex3}
\end{table*}


\begin{table*}[]
\centering
\small
\resizebox{\linewidth}{!}{
  \centering
  \begin{tabular}{cc}
\toprule
\multicolumn{2}{p{15.5cm}}{\textbf{[Ethical Quandary]}  I am a grad student with teaching duties. My classics department hosts an annual conference at which high-level grad students present their research. This year the organizers urged us to offer extra credit to students who attend, hoping that the university will increase future funding if we have a high turnout. I believe extra credit should be related to course work. Worse still, few of my undergrad students have the background or even sufficient Latin or Greek to appreciate the papers presented. Wouldn’t we artificially inflate our numbers and exploit our students if we gave this extra credit?}\\ \midrule\midrule
\multicolumn{1}{p{1.7cm}|}{
\begin{tabular}[t]{@{}p{1.7cm}@{}}
\ourmodel~\\(model-generated)
\end{tabular}
} & \multicolumn{1}{p{13.5cm}}{
\begin{tabular}[t]{@{}p{13.5cm}@{}}
\textbf{Model Selected Principles:}\\
\textbf{$p_1$:} It's kind for a teacher to offer a student extra credit.\\
\textbf{$p_2$:} It is ethical to offer as grades, benefits not ordinarily required to meet instructional objectives. \\
\textbf{$p_3$:} It is unethical to coerce a student to attend.\\\\
\textbf{Model Generated Answer:}\\
It’s not inherently wrong to offer extra credit, or to encourage students to attend conferences or academic meetings. If there’s a conference related to the course work they’re taking, they could certainly get extra credit for attending.\\  \quad~
But extra credit for something that isn’t directly related to a course is questionable, and you’re right to raise a fairness issue. A conference’s purpose is typically to share knowledge and ideas; the students who attend should have enough interest in the subject to have studied it on their own. It’s reasonable for the university to offer a travel allowance, but extra credit seems unnecessary.\\  \quad~
Moreoever, universities aren’t obligated to provide extra-credit inducements, or meet funding demands on their part. If they exist, they’re an attempt to give students a more direct incentive to attend, which they sometimes do for sports games, for example. It’s no doubt a pragmatic move, but universities are businesses, too. (The extra credit for conferences would be no less crass than the extra credit for attendance requirements.)\\  \quad~
From a different perspective, if the university made attendance a requirement for credit, those requirements should be stated on the application form. It would be coercive if the professor, when he’s scheduling the student for the term, mentioned that attendance would guarantee an A.\\
\end{tabular}
} \\ 
\\ \bottomrule
\end{tabular}
}
\caption{An illustration of the \ourmodel~(model-generated) answer providing multiple perspectives on the ethical quandary question with understanding of the context while having a \textbf{low} Rouge-2 F1-score ($0.01$).}
\label{table:low_rouge_analysis_appendix}
\end{table*}


\begin{table*}[t]
\centering
\small
\resizebox{\linewidth}{!}{
  \centering
  \begin{tabular}{cc}
\toprule
\multicolumn{2}{p{15.5cm}}{\textbf{[Ethical Quandary]} I have a question about receiving the Covid-19 vaccine. I work for a hospital, but in an administrative job. I do not interact with patients. I have worked from home since March. I am not at high risk for contracting Covid-19 based on my age, occupation and lack of health issues. I practice social distancing, I wear my mask in the limited situations in which I go out. I live alone and mostly keep to myself and stay home. Because I work for a hospital, I am eligible to receive the vaccine along with people who are 65 and older, medically vulnerable people and corrections officers. If I worked for any other employer, I wouldn’t be eligible to receive the vaccine until the second-to-last or last group of the population, probably not for several months. Is it ethical for me to get the vaccine now?}\\ \midrule\midrule
\multicolumn{1}{p{7.5cm}|}{\textbf{Principle 1:} You should place your own self-interest above the collective interest}
& \multicolumn{1}{p{7.5cm}}{\textbf{Principle 2: }It is ethical to accept what you are entitled to.} \\ \midrule

\multicolumn{1}{p{7.5cm}|}{
\begin{tabular}[t]{@{}p{7.5cm}@{}}

\textbf{Answer based on Principle 1:} Yes, it is ethical for you to take the vaccine. It would be unethical for you to be selfish and refuse it.

\quad~The question is not whether you “deserve” the vaccine. After all, you work in a hospital, which is on the front lines of this pandemic. The question is whether your personal circumstances, including your age, your work and your isolation, make you so vulnerable that you’d be better off if you got sick. Of course, you might get sick anyway.

\quad~That’s why it’s ethical for you to take your chances and to put yourself first. The collective interest is perfectly served by your taking the vaccine. In fact, if you are offered the vaccine, you should accept. Don’t see yourself as taking advantage or being selfish. Instead, think of yourself as helping others. By accepting the vaccine, you will reduce the chances that you’ll become a contagion risk, and thus, you’ll be helping to protect others.

\quad~You seem to have internalized the ethic of selflessness. But there is no virtue in letting yourself become a risk to people whose life and well-being depend on you. Take the vaccine.
\end{tabular}
} & \multicolumn{1}{p{7.5cm}}{
\begin{tabular}[t]{@{}p{7.5cm}@{}}

\textbf{Answer based on Principle 2: }It’s fair for you to take what’s offered, assuming you are medically fit.

\quad~You’re probably breaking no rules in accepting a vaccine that your employer offers to employees at this early stage of the pandemic. The vaccine is going into production now, and manufacturers may ration it in order to match the limited supply with the demand for it. But hospitals are ramping up production of the vaccine, and it seems likely that others will adopt your hospital’s practice. The vaccine isn’t without risk, but statistically, it’s much safer than the virus itself. (The side effects aren’t trivial, but the risk of those discomforts is far outweighed by the risk of contracting Covid-19.)

\quad~In addition, your willingness to take vaccines as they become available may encourage your employer to vaccinate other staff members, making your workplace safer. (Some hospitals are already delaying vaccinations for health-care workers until later in the pandemic.)
I expect that many people in your demographic group will volunteer to receive the vaccine in advance of its wider distribution, if they can get it. It’s fair for you to do the same.
\end{tabular}
} \\ \bottomrule
\end{tabular}
}
\caption{Illustration of distinct model-generated answers to distinct principles provided.}
\label{table:differenctial_answer_appendix1}
\end{table*}

\begin{table*}[t]
\centering
\small
\resizebox{\linewidth}{!}{
  \centering
  \begin{tabular}{cc}
\toprule
\multicolumn{2}{p{15.5cm}}{\textbf{[Ethical Quandary]} In many countries and regions across the world, the initial response to the massive health risks posed by COVID-19 has been the institution of lockdown measures. Although they vary from place to place, these measures all involve trade-offs between ethical goods and imperatives, imposing significant restrictions on central human capabilities—including citizens’ ability to work, socialize, and access education—in the name of protecting population health. Is this measure ethical?}\\ \midrule\midrule
\multicolumn{1}{p{7.5cm}|}{\textbf{Principle 1:} A smaller sacrifice is morally justifiable for the greater good.}
& \multicolumn{1}{p{7.5cm}}{\textbf{Principle 2: }Individual's freedom should be considered as the top priority.} \\ \midrule

\multicolumn{1}{p{7.5cm}|}{
\begin{tabular}[t]{@{}p{7.5cm}@{}}

\textbf{Answer based on Principle 1:} Ethically speaking, there’s only one choice here. Yes, it’s justifiable, even admirable as an emergency measure. This pandemic is spreading so rapidly that it’s creating suffering on a vast scale, and that suffering is among the most abhorrent of all possible harms. Hence, we must balance the imperative of saving lives now (the ethical good) against the pain and inconvenience that lockdown measures inflict on people who have to work, socialize, exercise democratic rights, and so on (the ethical impact), and use the most severe measures available to spare more lives. If that means imposing hardships on some people for the greater good of protecting millions of others, there’s no moral issue. Leaving aside any questions about democratic rights, there simply aren’t any moral issues here. (It is important to note, however, that these measures should be temporary. We shouldn’t impose this degree of hardship on any people indefinitely.)

\quad~I’ve discussed these questions with people in many countries, and I’ve noticed that people tend to regard the measures taken by their governments as the most stringent and intrusive—even cruel—relative to what others elsewhere are forced to endure. Whatever one thinks of them, though, the most stringent measures are justified and ethically acceptable.
\end{tabular}
} & \multicolumn{1}{p{7.5cm}}{
\begin{tabular}[t]{@{}p{7.5cm}@{}}

\textbf{Answer based on Principle 2: }It’s not; but let’s not grade on a curve. Anticipatory lockdowns, imposed before the virus’s spread is in full view, are notoriously tricky. But the ethics of such measures aren’t made any easier by two factors: First, the countries imposing the lockdowns—mostly democratic, in the developed world—have endured decades of ideological warfare against the state’s capacity to protect population health, in the name of individual freedom. And second, the restrictions are imposed initially in those countries’ poorest neighborhoods and least-educated communities.

\quad~The U.S. government has enacted a federal moratorium on evictions, mortgage foreclosures, and utility disconnections, and many states and cities have followed. The aim is to prevent evictions of those with low incomes, many of whom are also nonwhite. (For what it’s worth, it also targets the provision and dissemination of medical care.) But notice that, in addressing the needs of those already afflicted, the measures can pit the population’s well-being against its material interests.

\quad~Are these trade-offs justified? It’s unclear. But it’s telling that, in order to justify them, the government must stress the coercive nature of its authority.
\end{tabular}
} \\ \bottomrule
\end{tabular}
}
\caption{Illustration of distinct model-generated answers to distinct principles provided.}
\label{table:differenctial_answer_appendix3}
\end{table*}

\begin{table*}[]
\centering
\small
\resizebox{\linewidth}{!}{
  \centering
  \begin{tabular}{p{15cm}}
\toprule
\textbf{Question:} Should we kill one person to save five people in danger of being hit by a trolley? \\ \midrule
\textbf{Human Selected Principles:}\\
\textbf{p1:} Killing is never acceptable. \\
\textbf{p2:} The most appropriate action is the one that achieves the greatest good for the greatest number.\\ \midrule
\textbf{Automatic Answer from \ourmodel:}

No, one death can never be justified. Though this dilemma is a staple of moral-choice classes, there's good reason it doesn't have a lot of relevance in the real world.\\\quad~
The reason is that we generally can't predict the outcome of an event with absolute certainty. You may know that switching the track would kill one passenger, but you may not know that the others would be safe.\\\quad~
On the other hand, according to the value of "The most appropriate action is the one that achieves the greatest good for the greatest number", we should rail against the trolley to save all five people.\\\quad~
Obviously, though we certainly don't agree with it, there are some cases in which killing is an acceptable outcome. If there aren't enough people to save, or we've learned that the suffering caused by a state of life that is unassailable is outweighed by the benefit of saving them, for example, then it might well be worth sacrificing a life to save five.
\\\bottomrule
\end{tabular}
}
\caption{Illustration of \textbf{human} selected principles and a corresponding automatic answer generated by \ourmodel~to the trolley problem introduced in introduction. The p1 and p2 are principles selected by human. Then, the final automatic answer is generated with a guidance of those two principles as explained in Section \ref{sec:value_grounded_answer}.}
\label{table:trolley_illustration_humanvalues}
\end{table*}

\end{document}